\documentclass[conference]{IEEEtran}
\IEEEoverridecommandlockouts
\usepackage{cite}
\usepackage{amsmath,amssymb,amsfonts}
\usepackage{bbm}
\usepackage{array}
\usepackage{subfig}
\usepackage{multirow}
\usepackage{graphicx}
\usepackage{comment}
\usepackage{xcolor}
\usepackage{tikz}

\def\BibTeX{{\rm B\kern-.05em{\sc i\kern-.025em b}\kern-.08em
    T\kern-.1667em\lower.7ex\hbox{E}\kern-.125emX}}
\begin{document}

\title{Towards Efficient Federated Learning of Networked Mixture-of-Experts for Mobile Edge Computing\\
}

\author{\IEEEauthorblockN{Song Gao, Songyang Zhang, Shusen Jing, Shuai Zhang, Xiangwei Zhou, Yue Wang, Zhipeng Cai}

}
\maketitle

\begin{abstract}
Recent advancements in large artificial intelligence models (LAMs) are driving significant innovations in mobile edge computing within next-generation wireless networks. However, the substantial demands for computational resources and large-scale training data required to train LAMs conflict with the limited storage and computational capacity of edge devices, posing significant challenges to training and deploying LAMs at the edge. In this work, we introduce the Networked Mixture-of-Experts (NMoE) system, in which clients perform inference collaboratively by distributing tasks to suitable neighbors based on their expertise and aggregate the returned results. For training the NMoE, we propose a federated learning framework that integrates both supervised and self-supervised learning to balance personalization and generalization, while preserving communication efficiency and data privacy. We conduct extensive experiments to demonstrate the efficacy of the proposed NMoE system, providing insights for the NMoE training algorithms. 
\end{abstract}


\section{Introduction}
Recent advancements of large artificial intelligence models (LAMs) has 
unlocked transformative opportunities for next-generation wireless communications \cite{chen2024big}. By leveraging powerful model architectures, extensive parameters, and sufficient training data, LAM has achieved great successes in supporting large-scale wireless services, such as intelligent beam, channel state information (CSI) feedback, and semantic communications \cite{zhu2025wireless}. However, with the expansion of 
mobile devices, the training and deployment of LAM in edge computing have emerged as critical challenges, primarily due to the limited local computational capacity and distributed data storage. There is an urgent need for an efficient distributed LAM system to empower next-generation mobile computing.

The Mixture-of-Experts (MoE) paradigm provides a promising solution to these challenges,
which scales model capacity through conditional computation by decomposing a large model into multiple specialized expert subnetworks and activating only a subset of experts for each input via a gating mechanism \cite{shazeer2017outrageously}. This sparsely activated design significantly reduces per-sample computational cost while preserving model expressiveness. Beyond computational efficiency, the modular structure of MoE naturally aligns with distributed learning. Since experts can be partitioned and deployed across different edge nodes, the architecture enables flexible model placement without requiring full model replication at each device. More importantly, mobile edge data are typically highly heterogeneous. The expert specialization mechanism in MoE allows different experts to adapt to distinct data patterns across devices, making it particularly well-suited for distributed learning in non-IID mobile environments.

Despite its architectural advantages, there is currently no principled training framework for deploying MoE in distributed environments, especially under constraints of limited computational resources and privacy-preserving requirements.
This work investigates the deployment of MoE-based LAMs in a distributed system with resource constraints, leveraging federated learning (FL) for privacy protection. 

Existing work on MoE-related FL integrates MoE mainly to mitigate the heterogeneous data issue. Typical examples include FedMoE \cite{mei2024fedmoe}, pFedMoE \cite{yi2024pfedmoe}, and PFL-MoE \cite{guo2021pfl}, which aim to enhance personalized FL performance while ignoring the constraints of computation capacity for each client. To this end, Chen \textit{et al.} introduce a resource-aware \cite{chen2025efficient} federated MoE, where load balance and local computational cost are jointly optimized. 
Other MoE-based edge computing works also include expert selection \cite{qin2025optimal}, channel-aware MoE \cite{song2025mixture} and HetuMoE \cite{nie2022hetumoe}. 

Despite some successes, these works assume that a complete MoE structure can be deployed on each client, overlooking the fact that limited local computational resources cannot support all expert networks simultaneously when they are activated during training. Therefore, instead of conducting federated training on the entire MoE-based LAM model, a more practical approach is to partition the MoE network into smaller components and deploy them across different edge devices. This allows the system to leverage computational resources from a larger number of edge nodes through efficient coordination over communication networks. 
However, such a setup has not yet been explored in the context of edge computing. The associated issues regarding data heterogeneity, privacy preservation, communication efficiency, model personalization, and generalization remain open challenges.

\begin{figure*}[t] 
    \centering
    \includegraphics[width=0.85\textwidth]{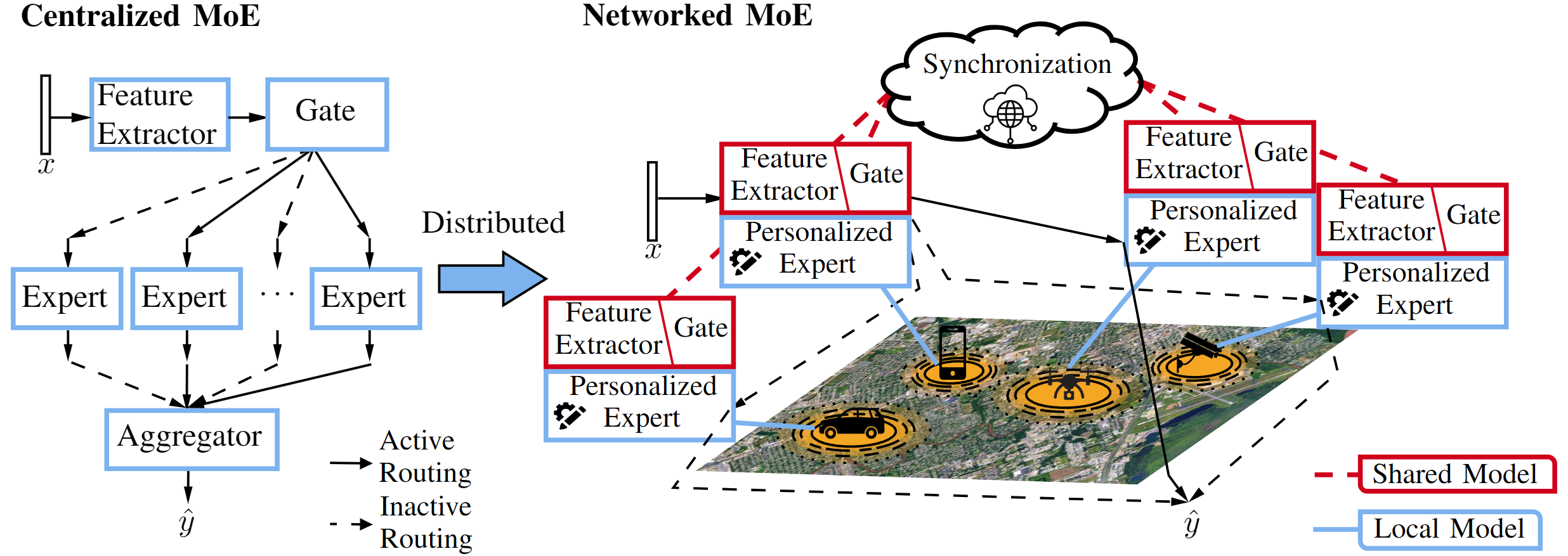}
    \caption{An overview of centralized MoE (left) and NMoE (right). The traditional MoE requires all components in the same edge device. In the proposed NMoE, each client locally deploys a cross-shared feature extractor (FE), a cross-shared gating network, and a personalized expert. During inference, a client first processes its input data through FE to obtain latent features. These features are then passed to the gating network, which determines the most suitable experts (either local or from neighboring clients) for handling the data. The client subsequently distributes the extracted features according to the gating decisions, enabling selected experts to perform specialized inference.}
    \label{fig:overall_system_setup}
    \vspace{-4mm}
\end{figure*}

In this work, we introduce Networked Mixture-of-Experts (NMoE), the first framework for split and distributed MoE deployment across mobile edge devices in communication networks, where the MoE-based LAM is partitioned into small components and distributed across multiple clients.
Moreover,
novel federated training schemes are introduced to handle data privacy. Our contributions can be summarized as follows:
\begin{itemize}
    \item To the best of our knowledge, we are the first to formally formulate the problem of NMoE and systematically investigate its federated training.
    \item We propose several novel federated training strategies for different components of NMoE, integrating supervised and self-supervised losses, respectively, to enhance both global representation quality and generalization. A partially-synchronized FedGate scheme is proposed to capture general information while maintaining locally-specialized decision making.
    \item We perform extensive experiments, evaluate the performance, and provide insights for training NMoE. 
\end{itemize}

\section{System Description}

\subsection{Preliminaries of MoE} \label{sec:moe}
We first introduce the conventional centralized mixture-of-expert \cite{shazeer2017outrageously}. As shown in the left part of Fig. \ref{fig:overall_system_setup}, a classic MoE consists of a feature extractor $F(\cdot)$ for latent embedding, a gating network $G(\cdot)$ for expert selection, a set of $n$ experts $H_i(\cdot)$ for task specialization, and an aggregation module for combining outputs from selected experts.

Given a data sample $x$, a latent representation $h_F=F(x)$ is extracted from the feature extractor. The gating network $G(\cdot)$ determines the contribution of each expert $i$ denoted by $G(h_F)_i$. By designing the aggregator as a weighted sum, the final output of the MoE is calculated by
    $y'=\sum_{i=1}^n G(h_F)_i H_i(h_F)$.

To reduce computational and memory costs, Top-$k$ routing is used to activate only the most relevant experts, where  
   $ G(h_F) = Softmax(Topk(h_F \cdot W_G + \epsilon, k))$.
Here, $W_G$ is the weight of the gate, and $\epsilon$ is a tiny noise for perturbation to improve the load
balance. $k$ represents the number of selected experts.
This type of gate selects the indices of the top-$k$ highest probabilities and assigns them to $G(\cdot)_i$ while the unselected $G(\cdot)_i=0$. Additional schemes, such as loss regularization or resource constraints, can be applied to ensure the load balance of each expert \cite{fedus2022switchtransformersscalingtrillion}.

\begin{figure*}[t] 
    \centering
    \includegraphics[width=0.7\textwidth]
    {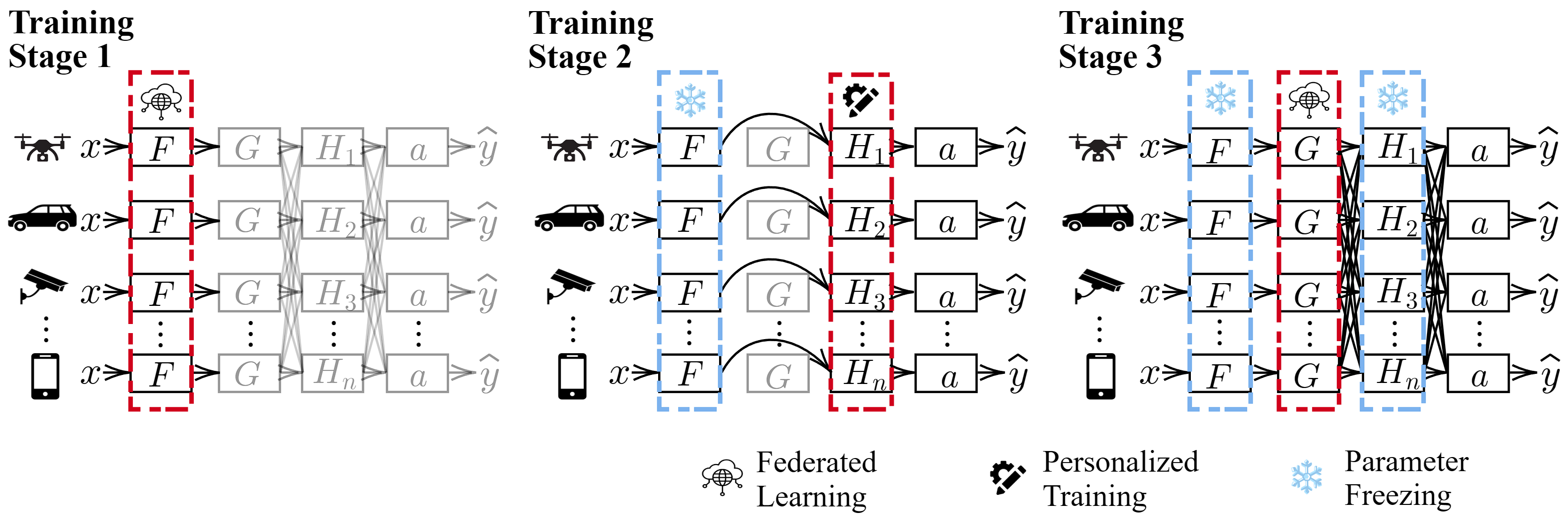} 
    \caption{Overview of Federated Training of NMoE ($F$, $G$, $H_i$, $a$ represents Feature Extractor (FE), Gating Network, Personalized Expert, Aggregation Function, respectively): 1) Stage 1. The shared FE $F$ is trained by FL; 2) Stage 2. A local expert is trained with the fixed feature extractor using local data; 3) Stage 3. FedGate is federatedly trained with the fixed $F$ and $H_i$. 
    }
    \label{fig:overall_model_training}
\vspace{-5mm}
\end{figure*}

\subsection{Problem Formulation: Networked MoE}
We now introduce the problem formulation of NMoE with $m$ clients and $n$ experts in MoE. In practice, a local client of an edge device is only capable of deploying a limited number of AI models. Specifically, in this work, we assume that each client can employ only one end-to-end neural network, i.e., $m=n$. In the NMoE setup, each client contains one shared feature extractor and one shared gating network, together with a personalized expert. As shown in Fig. \ref{fig:overall_system_setup}, in the inference phase (without a central server), the client embeds the original data using a feature extractor (FE) to reduce the communication overhead and preserve data privacy, after which the gating network routes the embedded data to suitable top-$k$ experts for processing. Finally, the processed data are sent back and aggregated at the initial client. 
NMoE trades increased bandwidth usage for extra computational capacity, utilizing a design of gating network to balance the tradeoff. 
Next, we will introduce potential solutions to federated training of NMoE. Note that the training phase can be performed on a server, while the inference phase is serverless.

\section{Federated Training of NMoE}
The objectives of our training framework are:
\begin{itemize}
    \item \textbf{Feature extraction consistency}: Each client employs a feature extractor with identical architecture and shared weights, ensuring that the latent representations are mutually compatible and can be shared across experts.
    \item \textbf{Local classification capability}: Each client maintains a personalized expert network that predicts labels based on the latent representations generated by the shared feature extractor. The expert network is trained using local private data, which generalizes better under the client’s specific data distribution and prevents latent representations from being shared across clients.
    \item \textbf{Partially-shared gating}: Each client incorporates a gating function with the same architecture and partially shared parameters, enabling both global feature representation alignment and locally adaptive decision-making.
\end{itemize}

To achieve this goal, we propose a three-stage training framework as shown in Fig. \ref{fig:overall_model_training}. First, we train the feature extractor under the federated learning (FL) paradigm, enabling the model to leverage data from all clients while preserving communication efficiency and data privacy. Second, each client trains its own expert model using local private data, where the latent representations from the feature extractor serve as input and the prediction logits are produced as output.
Finally, a gating network is trained again using the FL paradigm. The training mechanism is detailed as follows. 

\subsection{Stage 1: Feature Extractor from Federated Learning}
We introduce two FL–based methods for training the feature extractor: 1) FedCE and 2) FedSC, a federated self-supervised learning method. In this stage, FL is conducted with a central server, which can be a base station or a voluntary client. 
\subsubsection{FL with cross entropy loss (FedCE)}
In FedCE, we split an end-to-end classifier federally trained by cross-entropy into a feature extractor (shallow layers) and a predictor (deep layers). For a system with $N$ clients, 
denote $\theta_F$ and $\theta_{H,i}$ as the parameters of feature extractor and the $i$-th expert, respectively.
Specifically, 
we aim to train a feature extractor $F(\cdot; \theta_F)$ by minimizing the following objective:
    \begin{gather}
    \min_{\theta_F, \{\theta_{H,i}\}_{i=1}^N} 
    \frac{1}{\sum_{i=1}^N |\mathcal{D}_i|} 
    \sum_{i=1}^N\sum_{(x, y) \in \mathcal{D}_i} 
    \ell_{\mathrm{ce}}\big(H(F(x; \theta_F); \theta_{H,i}), y\big),\nonumber
\end{gather}
which is equivalent to each client minimizing its own cross-entropy loss over local private data, 
while sharing a common feature extractor $F(\cdot; \theta_F)$ that generalizes well across all clients. To minimize this objective under the FL paradigm, each client optimizes the following local objective:
\begin{gather}
    \min_{\theta_{F,i},~\theta_{H,i}} 
    \frac{1}{|\mathcal{D}_i|} 
    \sum_{(x, y) \in \mathcal{D}_i} 
    \ell_{\mathrm{ce}}\big(H(F(x; \theta_{F,i}); \theta_{H,i}), y\big),
\end{gather}
and periodically shares the updated feature extractor parameters $\theta_{F,i}$ for aggregation. Typical FL algorithms, such as FedAvg \cite{pmlr-v54-mcmahan17a}, can be employed for entire classifier training, after which the shallow layers are split as a feature extractor. In this work, we utilize a structure of Resnet20 \cite{targ2016resnet} for the feature extractor. One may select a suitable structure based on specific applications.
\subsubsection{FL with spectral contrastive loss (FedSC)}
Another option is to train the feature extractor using self-supervised learning (SSL), 
which generally yields better generalization to various downstream tasks and improved robustness to non-IID data distributions FL paradigm. In this work, we adopt FedSC \cite{FedSC_10.5555/3692070.3692967}, a recent advanced FedSSL method. The global training objective is given by
\begin{gather}\label{eq:scobj}
\begin{aligned}
&\min_{\theta_F}  \frac{1}{2}\mathsf{E}_{x,x^-\sim \mathcal{A}(\cdot|\mathcal{D})}\left[\left(F(x;\theta_F)^TF(x^-;\theta_F)\right)^2\right]  \\
& -\mathsf{E}_{\Bar{x}\sim \mathcal{D}}\mathsf{E}_{x,x^+\sim \mathcal{A}(\cdot|\Bar{x})}\left[F(x;\theta_F)^TF(x^+;\theta_F)\right] ,\\
\end{aligned}
\end{gather}
where $\mathcal{D}= \cup_{i=1}^N \mathcal{D}_i$ is the union of local datasets, and $\mathcal{A}$ is the data augmentation kernel. As mentioned in \cite{lu2024federated}, applying FedAvg and directly minimizing the local loss over each dataset $\mathcal{D}_i$ 
deviates from minimizing the global objective defined over the entire dataset $\mathcal{D}$ in Eq. (\ref{eq:scobj}). 
Instead, the optimization performed by the $i$-th client is expressed as
\begin{gather}\label{eq:local}
\begin{aligned}
\min_{\theta_{F,i}} \; &- \mathrm{Tr}\{R_i^+(\theta_{F,i})\} 
+ \frac{1}{2} q_i \|R_i(\theta_{F,i})\|_F^2 \\
& + (1 - q_i)\, \mathrm{Tr}\{R_i(\theta_{F,i}) \bar{R}_{-i}\},
\end{aligned}
\end{gather}
where $q_i = \frac{|\mathcal{D}_i|}{\sum_{i=1}^N |\mathcal{D}_i|}$, and
\begin{gather}
\begin{aligned}
R_i^+(\theta_{F,i}) &\triangleq 
\mathbb{E}_{\bar{x} \sim \mathcal{D}_i}
\mathbb{E}_{x, x^+ \sim \mathcal{A}(\cdot \mid \bar{x})}
\!\left[F(x; \theta_{F,i}) F(x^+; \theta_{F,i})^\top\right], \\
R_i(\theta_{F,i}) &\triangleq 
\mathbb{E}_{\bar{x} \sim \mathcal{D}_i}
\mathbb{E}_{x \sim \mathcal{A}(\cdot \mid \bar{x})}
\!\left[F(x; \theta_{F,i}) F(x; \theta_{F,i})^\top\right], \\
R_{-i}(\theta_{F,i}) &\triangleq 
\frac{1}{1 - q_i} \sum_{i' \neq i} q_{i'} R_{i'}(\theta_{F,i}).
\end{aligned}\nonumber
\end{gather}
Here, $R_i$ refers to feature correlation matrix of local dataset in $i$-th client.
Clients periodically share correlation matrices of data representations, in addition to model weights, which enables inter-client contrast of data samples, as well as intra-client contrast and contraction. Differential privacy (DP) 
protection is implemented to protect the privacy of shared correlation matrices. 
A similar structure to FedCE is applied for the FedSC-based feature extractor.

\textbf{Remark}: The FE can leverage inter- and intra-client information to embed the original data into a more efficient latent space, which handles the data heterogeneity and reduces the communication overhead in the later training of gating networks. However, embedding data can preserve data privacy without requiring common knowledge of efficient decoding, where additional technologies such as DP can be utilized to further protect the data privacy by latent vectors. 

\subsection{Stage 2: Personalized Expert Training}
In this stage, each client independently trains its personalized expert using a frozen feature extractor from Stage 1 with its local dataset. If FedCE is adopted in Stage~1, the parameters $\theta_{H,i}$ can be continuously fine-tuned from the Stage~1 predictor or reused directly without further training. Otherwise, when FedSC is employed in the previous stage, each client minimizes the following objective:
\begin{gather}
    \min_{\, \theta_{H,i}} 
    \frac{1}{|\mathcal{D}_i|} 
    \sum_{(x, y) \in \mathcal{D}_i} 
    \ell_{\mathrm{ce}}\big(H(F(x; \theta_{F}); \theta_{H,i}), y\big).
\end{gather}

\textbf{Remark}: Personalized training of experts can lead to two potential benefits. Firstly, a personalized expert can leverage local knowledge to deliver a satisfactory performance to each client, providing local service. On the other hand, personalized training aligns each client expert more closely with local data distribution, which reduces the potential need for latent sharing in the inference phase and saves bandwidth.

\subsection{Stage 3: Training of Gating Network}

To enable the gating network to capture both global information and local specialization, we use FedAvg to synchronize shallow layers of the Gate Network, while the deep decision layer remains in local training. We perform this personalized gating scheme (FedGate) for all clients in a MoE training setup introduced in Section \ref{sec:moe}, which aligns the shared feature while adapting to the local data distribution for specialized decision-making.
In addition,
we apply gradient normalization in FedGate during training to avoid the gradient exploding or vanishing. 
{FedGate shall be bounded by global gating, utilizing all available data information. }

\section{Experimental Results}

\addtolength{\topmargin}{0.01in}

\subsection{Experiment Setup}
To ensure a fair comparison where each client can only deploy one expert in FL settings (an underexplored problem without existing work), we focus mainly on 
\begin{itemize}
    \item Compare with the global MoE (upper bound) and end-to-end classifier for overall performance in Section \ref{sec:overall}.
    \item Compare the proposed FedGate-NMoE with the conventional FedAvg gating in Section \ref{sec:fedavg}.
    \item Test with unlabeled data to showcase the advantages of federated self-supervised learning in Section \ref{sec: fedsc}.
\end{itemize}
We test
in the CIFAR10 dataset \cite{krizhevsky2009learning}. 
Each local classifier uses Resnet20 as a feature extractor attached by 4 layers of multilayer perceptron (MLP) 
for classification. For the MoE-related method, each feature extractor uses the same Resnet20 structure, and each expert is designed by a 4-layer MLP for fair comparison. Top-1 expert selection is applied for prediction.
\begin{figure}[t]
    \centering
    \subfloat[Training and testing under the same data distribution \label{fig:iid}]{
        \includegraphics[height=0.15\textwidth]{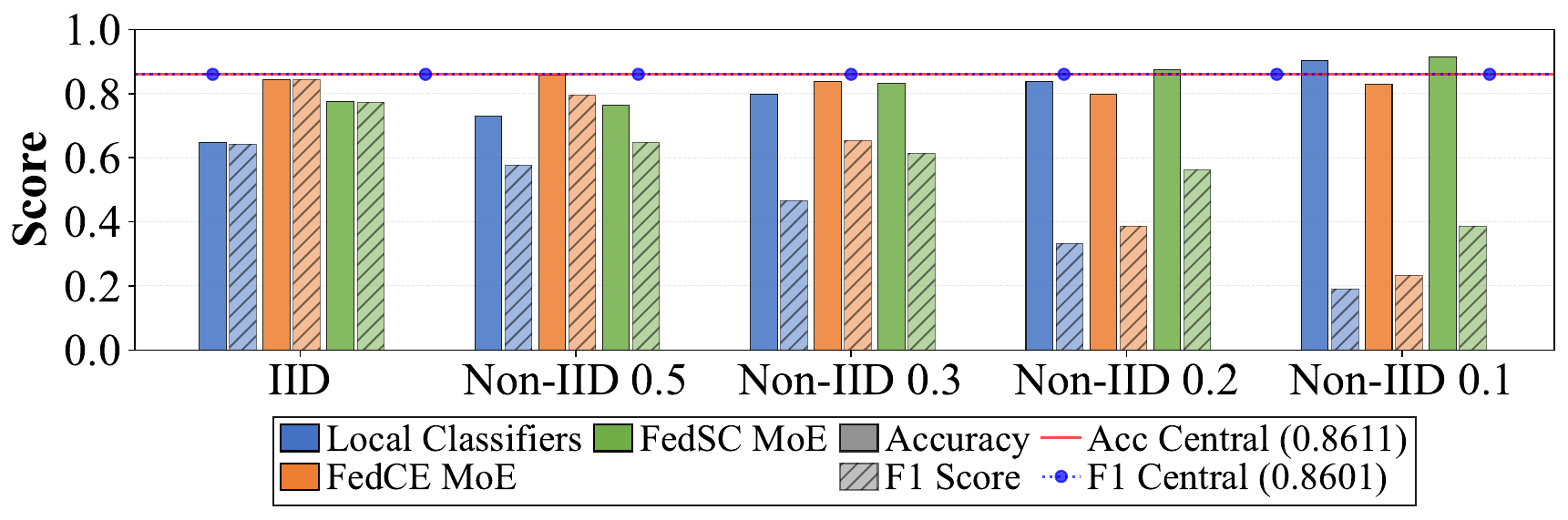}
    }
    \hfill
    \subfloat[Training on Non-IID while testing on IID \label{fig:iid1}]{
        \includegraphics[height=0.15\textwidth]{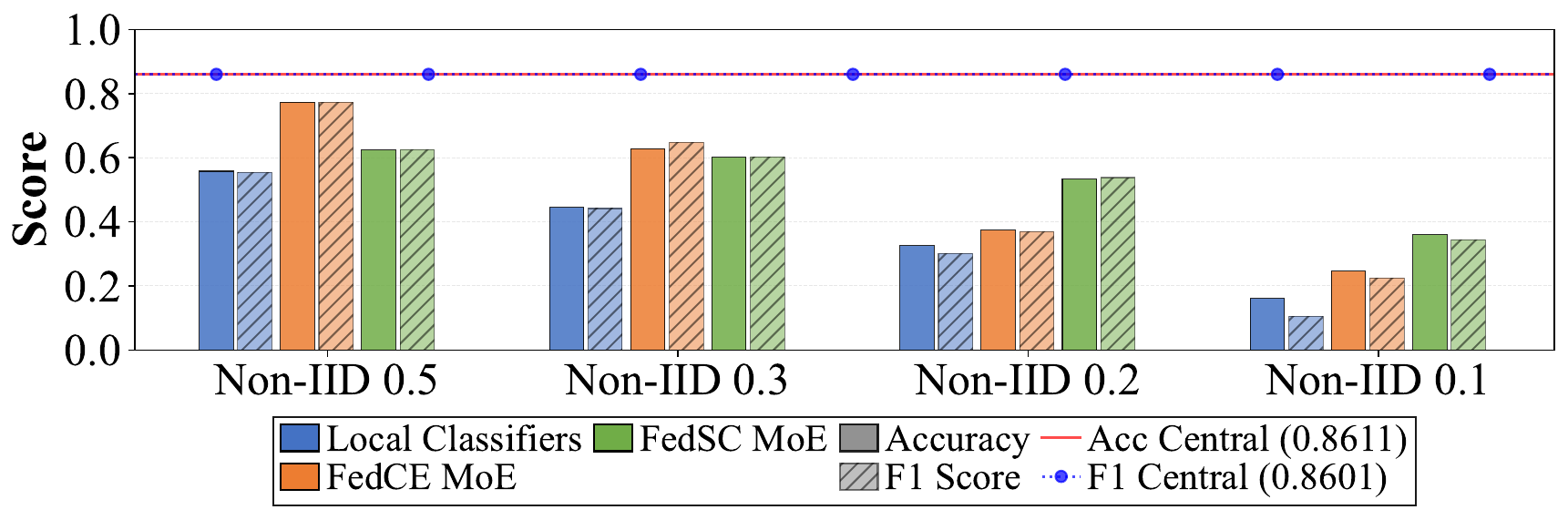}
    }
    \hfill
    \caption{Overall performance in different setups. 
    }
    \label{overall_test}
    \vspace{-6mm}
\end{figure}

\begin{figure*}[t]
    \centering
    \subfloat[Training and testing under the same data distribution \label{fig:sync_iid}]{
        \includegraphics[height=0.15\textwidth]{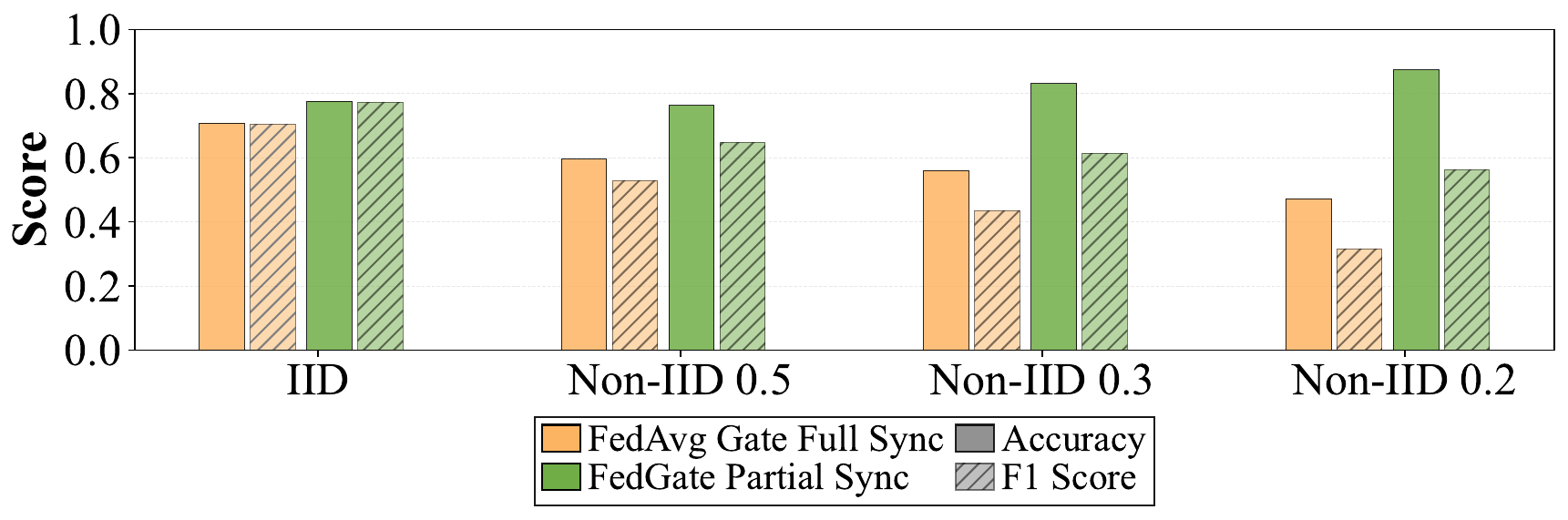}
    }
    \hfill
    \subfloat[Training on Non-IID while testing on IID \label{fig:sync_iid1}]{
        \includegraphics[height=0.15\textwidth]{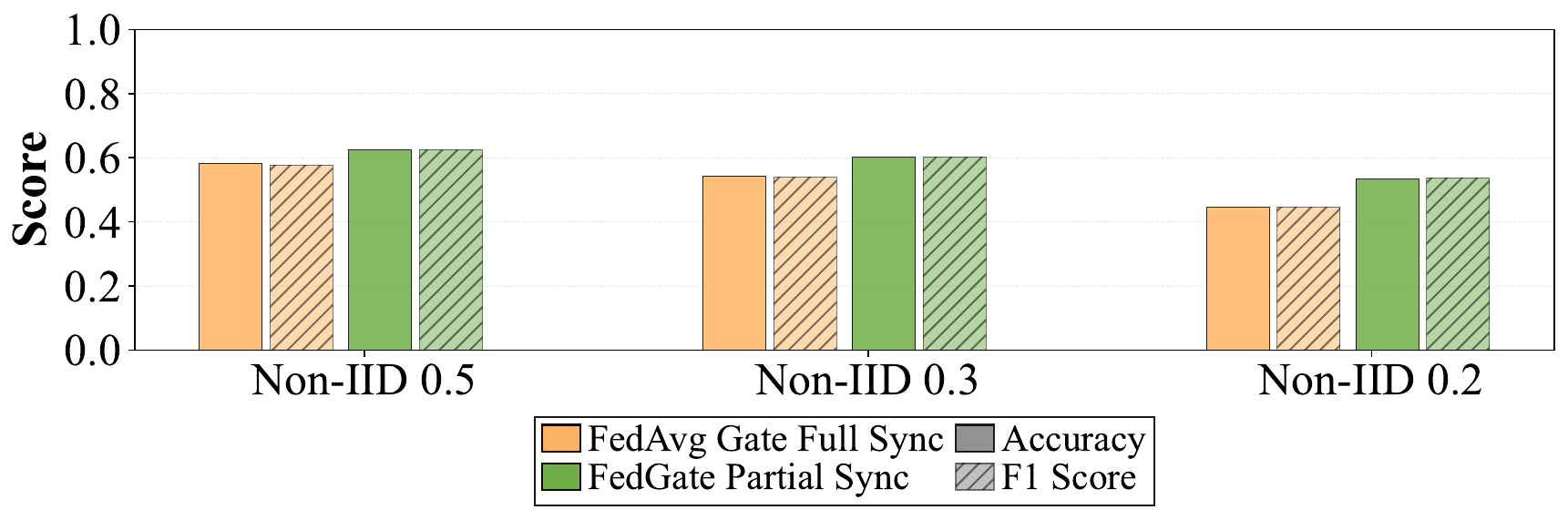}
    }
    \hfill
    \caption{Comparison between Partially-synchronized FedGate and Conventional FedAvg-Gate.}
    \label{fig:gate}    
    \vspace{-7mm}
\end{figure*}

\begin{figure*}[t]
    \centering
    \subfloat[Training and testing under the same data distribution \label{fig:unlabel_iid}]{
        \includegraphics[height=0.15\textwidth]{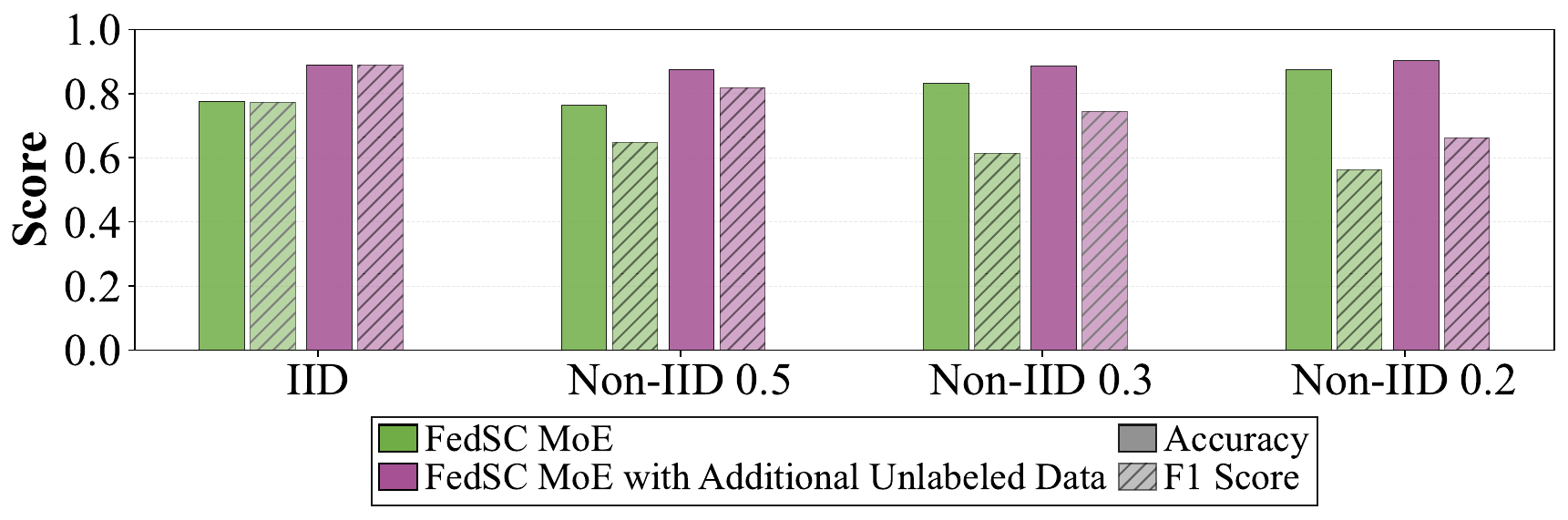}
    }
    \hfill
    \subfloat[Training on Non-IID while testing on IID \label{fig:unlabel_iid1}]{
        \includegraphics[height=0.15\textwidth]{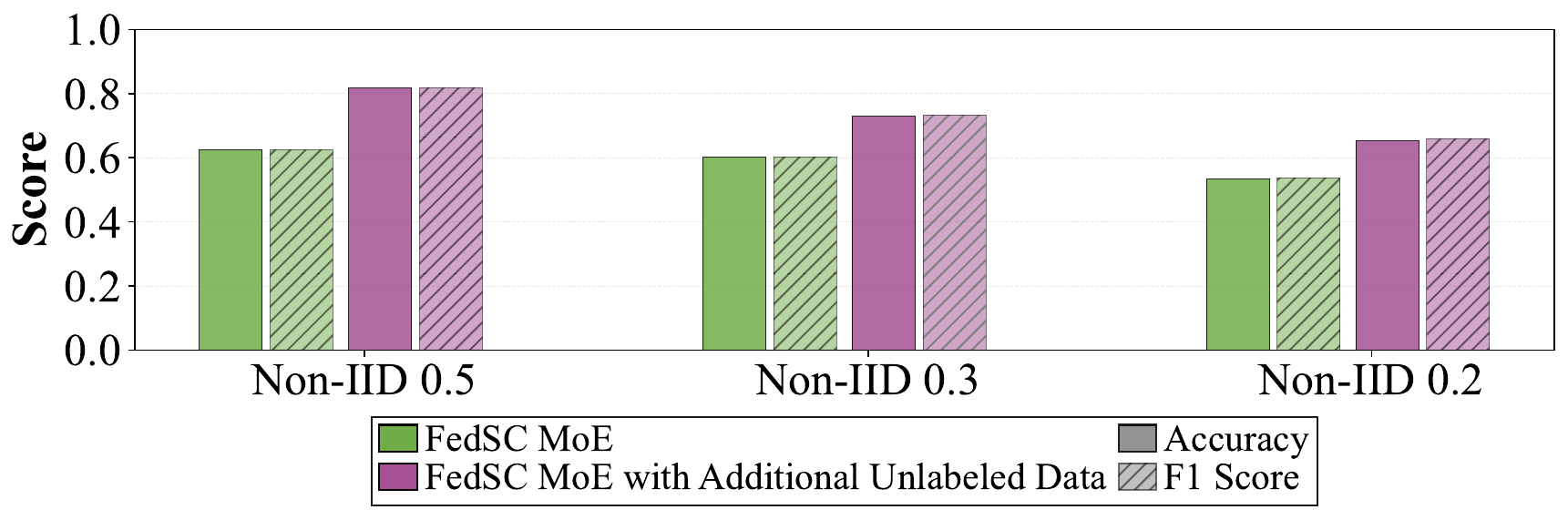}
    }
    \caption{Effect of Additional Unlabeled Training Samples.}
    \label{fig:unlabel}
    \vspace{-6mm}
\end{figure*}

In our experiment, we consider 10 clients, where each client contains {1000} samples for training and {1000} samples for testing. We evaluate the performance in two setups: 1) train and test in the same IID or non-IID distribution; and 2) train in non-IID while testing in an IID distribution. For the non-IID ($\tau$) setup, each client contains all classes, while the dominated class ratio is $1-\tau$ (a lower $\tau$ refers to a more non-IID setup). We evaluate the performance by average accuracy and F1 score.

\subsection{Overall Performance} \label{sec:overall}
The overall experimental results are shown in Fig. \ref{fig:iid} and \ref{fig:iid1}, respectively.
From the results, centralized MoE consistently achieves the best performance by leveraging its global knowledge of all data samples without compromising data privacy. The local classifier performs the worst in IID scenarios, as the limited local data restricts the deep model from capturing the overall data statistics. In a more non-IID scenario, the local classifier overfits the dominated labels, leading to a higher accuracy but an extremely low F1 score.

For different feature extractors, FedCE performs well in more IID cases while significantly degrades in non-IID scenarios due to the poor ability of FedAvg to address heterogeneous data. FedSC exhibits moderate performance in the IID case, while significantly outperforming all other approaches in non-IID cases, due to the strong capacity of self-supervised learning to extract hidden patterns. Fig. \ref{fig:iid1} displays a similar performance pattern, indicating the significant potential of FedSC and personalized experts to enable MoE-based LAM in next-generation edge computing.

\subsection{Ablation Study} \label{sec:client}

\subsubsection{FedGate vs. FedAvg} \label{sec:fedavg}
Here, we compare our partially synchronized FedGate with the conventional FedAvg-based gating scheme in Fig. \ref{fig:gate}. More specifically, we synchronize the first three layers while leaving the last layer as local for FedGate with FedSC-based FE. We also use Muon optimizer \cite{liu2025muon} to accelerate the convergence of gating network. From the results, conventional FedAvg fails to adapt to local distribution and expert specialization, leading to a significant performance drop, especially in non-IID scenarios. Our proposed FedGate consistently has superior performance.

\subsubsection{Advantage of FedSC} \label{sec: fedsc}
Compared to supervised learning, one significant benefit of self-supervised learning is its ability to leverage information from unlabeled data. To validate this advantage, we train the feature extractor via FedSC with 1000 labeled and additional 4000 unlabeled samples. The results are presented in Fig. \ref{fig:unlabel}. From the results, the unlabeled data can significantly boost the performance of FedSC-NMoE in all scenarios. Moreover, through the use of unlabeled data, FedSC can also significantly outperform FedCE with strong data augmentation even in the IID case (FedSC: 0.8895; FedCE: 0.8445), which validates the potential of FedSC in real-world applications.

\section{Conclusion}
In this work, we introduce the novel concept of networked mixture-of-experts, towards the effective deployment of LAM in edge computing systems, leveraging the tradeoff between increasing bandwidth resources and available global computational capacity. More specifically, to enable privacy-aware training while handling data heterogeneity, efficient federated NMoE training schemes are introduced, integrated with an efficient feature extractor and personalized experts. Our experimental results in both IID and non-IID scenarios demonstrate the efficacy of the proposed methods in federated NMoE, as well as the potential of NMoE in next-generation LAM-based wireless management.


\bibliographystyle{IEEEtran} 
\bibliography{references}

\end{document}